\documentclass[11pt,a4paper]{article}
\usepackage{url}
\usepackage[hyperref]{emnlp2018}

\usepackage{amsmath}
\usepackage{amssymb}
\usepackage{amsfonts}
\usepackage{array}
\usepackage{bm}
\usepackage{booktabs}
\usepackage[labelfont=bf,skip=5pt]{caption}
\usepackage{cases}
\usepackage{color}
\usepackage{diagbox}
\usepackage{enumitem}
\usepackage{graphicx}
\usepackage{latexsym}
\usepackage{makecell}
\usepackage[mathscr]{euscript}
\usepackage{mathtools}
\usepackage{multirow}
\usepackage{paralist}
\usepackage{sidecap}
\usepackage{soul}
\usepackage{subcaption}
\usepackage{tabu}
\usepackage{tabulary}
\usepackage{times}
\usepackage{url}
\usepackage{verbatim}

\definecolor{orange}{rgb}{1,0.5,0}
\definecolor{mdgreen}{rgb}{0.05,0.6,0.05}
\definecolor{mdblue}{rgb}{0,0,0.7}
\definecolor{dkblue}{rgb}{0,0,0.5}
\definecolor{dkgray}{rgb}{0.3,0.3,0.3}
\definecolor{slate}{rgb}{0.25,0.25,0.4}
\definecolor{gray}{rgb}{0.5,0.5,0.5}
\definecolor{ltgray}{rgb}{0.7,0.7,0.7}
\definecolor{purple}{rgb}{0.7,0,1.0}
\definecolor{lavender}{rgb}{0.65,0.55,1.0}

\aclfinalcopy
\newcommand{\ignore}[1]{}

\newcommand{\com}[1]{}

\newcommand{\term}[1]{\textbf{#1}} 
\newcommand{\sect}[1]{(\S\ref{#1})}

\newcommand{\task}[1]{{T}_{{#1}}}

\newcommand{\seq}[1]{\mathbf{#1}}
\newcommand{\spemb}{\seq{v}}
\newcommand{\hidden}{\seq{h}}

\newcommand{\tuple}[1]{\langle #1 \rangle}

\newcommand{\interalia}[1]{\citep[\emph{inter alia}]{#1}}
\newcommand{\argmax}[1]{\underset{#1}{\operatorname{arg}\,\operatorname{max}}\;}

\newcommand{\softmax}[1]{\underset{#1}{\operatorname{softmax}}\;}

\newcommand{\ind}[1]{\mathbb{I}(#1)}

\newcommand{\fnrole}[1]{\textsc{#1}}
\newcommand{\rolesForF}{\mathcal{Y}_f}
\usepackage{xspace}
\newcommand{\nullLabel}{\fnrole{null}\xspace}

\newcommand{\ai}{\texttt{AllenNLP}}

\newcolumntype{L}[1]{>{\raggedright\let\newline\\\arraybackslash\hspace{0pt}}m{#1}}
\newcolumntype{C}[1]{>{\centering\let\newline\\\arraybackslash\hspace{0pt}}m{#1}}
\newcolumntype{R}[1]{>{\raggedleft\let\newline\\\arraybackslash\hspace{0pt}}m{#1}}

\setul{1pt}{.4pt}

\usepackage{txfonts}
\newcommand{\fifthsuit}{\varheartsuit}

\title{Syntactic Scaffolds for Semantic Structures}

\author{ 
    Swabha Swayamdipta$^\spadesuit$ \quad
	Sam Thomson$^\spadesuit$  \\ 
	\bf Kenton Lee$^\diamondsuit$ \quad
	Luke Zettlemoyer$^{\fifthsuit}$ \quad
	Chris Dyer$^\heartsuit$ \quad
	Noah A. Smith$^{\fifthsuit\clubsuit}$ \\\\
	$^\spadesuit$Language Technologies Institute, Carnegie Mellon University, Pittsburgh, PA, USA \\
	$^\diamondsuit$Google AI Language, Seattle WA, USA \\
	$^\fifthsuit$Paul G. Allen School of Computer Science \& Engineering, University of Washington, Seattle, WA, USA \\
	    $^\heartsuit$Google DeepMind, London, UK \\
	$^\clubsuit$Allen Institute for Artificial Intelligence, Seattle, WA, USA \\
{\tt \{swabha,sammthomson\}@cs.cmu.edu 
\{kentonl,cdyer\}@google.com} \\
	{\tt \{lsz,nasmith\}@cs.washington.edu}
}

\date{}

\begin{document}
	\maketitle
	
	\begin{abstract}
We introduce the \textbf{syntactic scaffold}, an approach to incorporating syntactic information into semantic tasks. 
Syntactic scaffolds avoid expensive syntactic processing at runtime, only making use of a treebank during training, through a multitask objective. 
We improve over strong baselines on PropBank semantics, frame semantics, and coreference resolution, achieving competitive performance on all three tasks.
\end{abstract}

\section{Introduction}
\label{sec:intro}

As algorithms for the semantic analysis of natural language sentences have developed, the role of \emph{syntax} has been repeatedly revisited.  
Linguistic theories have argued for a very tight integration of syntactic and semantic processing \citep{Steedman:00,Copestake:00}, and many systems have used syntactic dependency or phrase-based parsers as preprocessing for semantic analysis \citep{Gildea:02b,Punyakanok:08,Das:14}.
Meanwhile, some recent methods forgo explicit syntactic processing altogether \citep{Zhou:15,He:17,Lee:17,Peng:17}.

Because annotated training datasets for semantics will always be limited, we expect that syntax---which offers an incomplete but potentially useful view of semantic structure---will continue to offer useful inductive bias, encouraging semantic models toward better generalization.
We address the central question: is there a way for semantic analyzers to benefit from syntax without the computational cost of syntactic parsing?

We propose a multitask learning approach to incorporating syntactic information into learned representations of neural semantics models \sect{sec:scaffolding}.
Our approach, the \textbf{syntactic scaffold}, minimizes an auxiliary supervised loss function, derived from a syntactic treebank.
The goal is to steer the distributed, contextualized representations of words and spans toward accurate semantic \emph{and} syntactic labeling.
We avoid the cost of training or executing a full syntactic parser, and at test time (i.e., runtime in applications) the semantic analyzer has no additional cost over a syntax-free baseline.
Further, the method does not assume that the syntactic treebank overlaps the dataset for the primary task.

\begin{figure*}
\begin{center}
  \includegraphics[width=.89\textwidth]{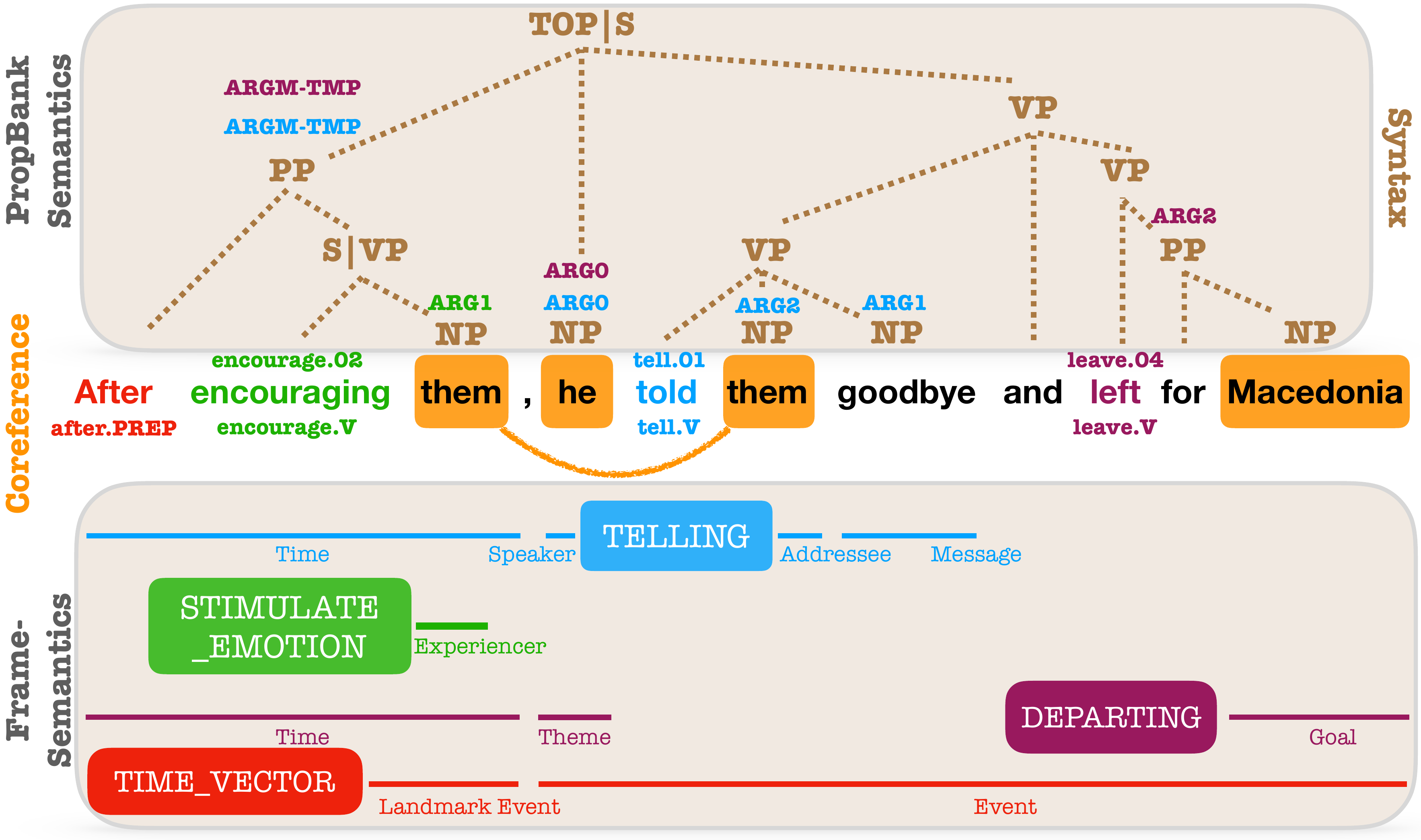}
\end{center}
\caption{
An example sentence with syntactic, PropBank and coreference annotations from OntoNotes, and author-annotated frame-semantic structures.
PropBank SRL arguments and coreference mentions are annotated on top of syntactic constituents.
All but one frame-semantic argument (\fnrole{Event}) is a syntactic constituent.
Targets evoke frames shown in the color-coded layers.
}
\label{fig:macedonia}
\end{figure*}

Many semantic tasks involve labeling spans, including semantic role labeling~(SRL; \citealp{Gildea:02a}) and coreference resolution~\cite{Ng:10} (tasks we consider in this paper), as well as named entity recognition and some reading comprehension and question answering tasks~\cite{Rajpurkar:16}.
These spans are usually syntactic constituents (cf.~PropBank;~\citealp{Palmer:05}), making phrase-based syntax a natural choice for a scaffold.
See Figure~\ref{fig:macedonia} for an example sentence with syntactic and semantic annotations.
Since the scaffold task is not an end in itself, we relax the syntactic parsing problem to a collection of independent span-level predictions, with no constraint that they form a valid parse tree.
This means we never need to run a syntactic parsing algorithm.

Our experiments demonstrate that the syntactic scaffold offers a substantial boost to state-of-the-art baselines for two SRL tasks \sect{sec:srl} and coreference resolution \sect{sec:coref}.
Our models use the strongest available neural network architectures for these tasks, integrating deep representation learning \citep{He:17} and structured prediction at the level of spans \citep{Kong:16}.
For SRL, the baseline itself is a novel globally normalized structured conditional random field, which outperforms the previous state of the art.\footnote{This excludes models initialized with deep, contextualized embeddings~\cite{Peters:18}, an approach orthogonal to ours.}
Syntactic scaffolds result in further improvements over prior work---3.6 absolute $F_1$ in FrameNet SRL, 1.1 absolute $F_1$ in PropBank SRL, and 0.6 $F_1$ in coreference resolution (averaged across three standard scores).
Our code is open source and available at \url{https://github.com/swabhs/scaffolding}.

\section{Syntactic Scaffolds}
\label{sec:scaffolding}

Multitask learning~\cite{Caruana:97} is a collection of techniques in which two or more tasks are learned from data with at least some parameters shared.  
We assume there is only one task about whose performance we are concerned, denoted $\task{1}$ (in this paper, $\task{1}$ is either SRL or coreference resolution).
We use the term ``scaffold'' to refer to a second task, $\task{2}$, that can be combined with $\task{1}$ during multitask learning.  
A scaffold task is \emph{only} used during training; it holds no intrinsic interest beyond biasing the learning of $\task{1}$, and after learning is completed, the scaffold is discarded.

A \textbf{syntactic scaffold} is a task designed to steer the (shared) model toward awareness of syntactic structure.  
It could be defined through a syntactic parser that shares some parameters with $\task{1}$'s model.  
Since syntactic parsing is costly, we use simpler syntactic prediction problems (discussed below) that do not produce whole trees.

As with multitask learning in general, we do not assume that the same data are annotated with outputs for $\task{1}$ and $\task{2}$.  
In this work, $\task{2}$ is defined using phrase-structure syntactic annotations from OntoNotes 5.0~\cite{Weischedel:13,Pradhan:13}.
We experiment with three settings: one where the corpus for $\task{2}$ does not overlap with the training datasets for $\task{1}$ (frame-SRL) and two where there is a complete overlap (PropBank SRL and coreference).  
Compared to approaches which require multiple output labels over the same data, we offer the major advantage of not requiring any assumptions about, or specification of, the relationship between $\task{1}$ and $\task{2}$ output.

\section{Related Work}
\label{sec:related}

We briefly contrast the syntactic scaffold with existing alternatives.

\paragraph{Pipelines.}
In a typical pipeline, $\task{1}$ and $\task{2}$ are separately trained, with the output of $\task{2}$ used to define the inputs to $\task{1}$~\cite{Wolpert:92}.
Using syntax as $\task{2}$ in a pipeline is perhaps the most common approach for semantic structure prediction ~\cite{Toutanova:08,Yang:17,Wiseman:16}.\footnote{
There has been some recent work on SRL which completely forgoes syntactic processing~\cite{Zhou:15}, however it has been shown that incorporating syntactic information still remains useful~\cite{He:17}.}
However, pipelines introduce the problem of cascading errors ($\task{2}$'s mistakes affect the performance, and perhaps the training, of $\task{1}$; ~\citealp{He:13}). 
To date, remedies to cascading errors are so computationally expensive as to be impractical (e.g., \citealp{Finkel:06}).
A syntactic scaffold is quite different from a pipeline since the output of $\task{2}$ is never explicitly used.

\paragraph{Latent variables.}
Another solution is to treat the output of $\task{2}$ as a (perhaps structured) latent variable. 
This approach obviates the need of supervision for $\task{2}$ and requires marginalization (or some approximation to it) in order to reason about the outputs of $\task{1}$.
Syntax as a latent variable for semantics was explored by \citet{Zettlemoyer:05} and \citet{Naradowsky:12}.
Apart from avoiding marginalization, the syntactic scaffold offers a way to use auxiliary syntactically-annotated data as direct supervision for $\task{2}$, and it need not overlap the $\task{1}$ training data.
\paragraph{Joint learning of syntax and semantics.}
The motivation behind joint learning of syntactic and semantic representations  is that any one task is helpful in predicting the other \cite{Lluis:08,Lluis:13,Henderson:13,Swayamdipta:16}.  This typically requires joint prediction of the outputs of $\task{1}$ and $\task{2}$, which tends to be computationally expensive at both training and test time.

\paragraph{Part of speech scaffolds.}
Similar to our work, there have been multitask models that use part-of-speech tagging as $\task{2}$, with
transition-based dependency parsing~\cite{Zhang:16} and CCG supertagging~\cite{Sogaard:16} as $\task{1}$.
Both of the above approaches assumed parallel input data and used both tasks as supervision.
Notably, we simplify our $\task{2}$, throwing away the structured aspects of syntactic parsing,
whereas part-of-speech tagging has very little structure to begin with.
While their approach results in improved token-level representations learned via supervision from POS tags, these must still be composed to obtain span representations. 
Instead, our approach learns span-level representations from phrase-type supervision directly, for semantic tasks. 
Additionally, these methods explore architectural variations in RNN layers for including supervision, whereas we focus on incorporating supervision with minimal changes to the baseline architecture.
To the best of our knowledge, such simplified syntactic scaffolds have not been tried before.

\paragraph{Word embeddings.}
Our definition of a scaffold task  \emph{almost} includes stand-alone methods for estimating word embeddings~\cite{Mikolov:13,Pennington:14,Peters:18}.  
After training word embeddings, the tasks implied by models like the skip-gram or ELMo's language model become irrelevant to the downstream use of the embeddings. 
A noteworthy difference is that, rather than pre-training, a scaffold is integrated directly into the training of $\task{1}$ through a multitask objective.

\paragraph{Multitask learning.}
Neural architectures have often yielded performance gains when trained for multiple tasks together~\cite{Collobert:11,Luong:15,Chen:17,Hashimoto:17}.
In particular, performance of semantic role labeling tasks improves when done jointly with other semantic tasks~\cite{Fitzgerald:15,Peng:17,Peng:18}.
Contemporaneously with this work, \citet{Hershcovich:18} proposed a multitask learning setting for universal syntactic dependencies and UCCA semantics \cite{Abend:13}.
Syntactic scaffolds focus on a primary semantic task, treating syntax as an auxillary, eventually forgettable prediction task.

\section{Syntactic Scaffold Model}
\label{sec:scaffolding-objective}

We assume two sources of supervision:  a corpus $\mathcal{D}_1$ with instances $\seq{x}$ annotated for the primary task's outputs $\seq{y}$ (semantic role labeling or coreference resolution), and a treebank $\mathcal{D}_2$ with sentences $\seq{x}$, each with a phrase-structure tree $\seq{z}$.

\subsection{Loss}
Each task has an associated loss, and we seek to minimize the combination of task losses,
\begin{align}
       \displaystyle \sum_{(\seq{x}, \seq{y}) \in \mathcal{D}_1}  \mathscr{L}_1(\seq{x}, \seq{y}) + 
 \delta \sum_{(\seq{x}, \seq{z}) \in \mathcal{D}_2}  \mathscr{L}_2(\seq{x}, \seq{z}) 
\end{align}
with respect to parameters, which are partially shared, where $\delta$ is a tunable hyperparameter.
In the rest of this section, we describe the scaffold task.
We define the primary tasks  in Sections \ref{sec:srl}--\ref{sec:coref}.

Each input is a sequence of tokens, $\seq{x} =  \langle x_{1}, x_2, \ldots, x_{n} \rangle$, for some $n$.
We refer to a \term{span} of contiguous tokens in the sentence as $\seq{x}_{i:j} = \langle x_i, x_{i+1}, \ldots, x_{j} \rangle$, for any $1 \leqslant i \leqslant j \leqslant n$.  
In our experiments we consider only spans up to a maximum length $D$, resulting in $O(nD)$ spans.

Supervision comes from a phrase-syntactic tree $\seq{z}$ for the sentence, comprising a syntactic category $z_{i:j} \in \mathcal{C}$ for every span $\seq{x}_{i:j}$ in $\seq{x}$ (many spans are given a
\nullLabel\ label).
We experiment with different sets of labels $\mathcal{C}$~\sect{sec:scaffold-tasks}.

In our model, every span $\seq{x}_{i:j}$ is represented by an embedding vector $\spemb_{i:j}$ (see details in \S\ref{sec:srl-input}).
A distribution over the category assigned to $z_{i:j}$ is derived from $\spemb_{i:j}$:
\begin{align}
p(z_{i:j} = c \mid \seq{x}_{i:j}) =&  ~\softmax{c}  \seq{w}_c \cdot \spemb_{i:j}
\end{align}
where $\seq{w}_c$ is a parameter vector associated with category $c$.
We sum the log loss terms for all the spans in a sentence to give its loss:
\begin{align}
 \mathscr{L}_2(\seq{x}, \seq{z}) &=  -\sum_{\substack{1 \leqslant i \leqslant j \leqslant n\\j-i \leqslant D}}  \log p(z_{i:j} \mid \seq{x}_{i:j}).
\end{align}

\subsection{Labels for the Syntactic Scaffold Task}
\label{sec:scaffold-tasks}

Different kinds of syntactic labels can be used for learning syntactically-aware span representations:
\begin{compactitem}
\item \textbf{Constituent identity}: $\mathcal{C} = \{0, 1\}$; is a span a constituent, or not?
\item \textbf{Non-terminal}: $c$ is the category of a span, including a \nullLabel\ for non-constituents.
\item \textbf{Non-terminal and parent}: $c$ is the category of a span, concatenated with the category of its immediate ancestor.
\nullLabel\ is used for non-constituents, and for empty ancestors.
\item \textbf{Common non-terminals}: Since a majority of semantic arguments and entity mentions are labeled with a small number of syntactic categories,\footnote{In the OntoNotes corpus, which includes both syntactic and semantic annotations, 44\% of semantic arguments are noun phrases and 13\% are prepositional phrases.} 
we experiment with a three-way classification among (i) noun phrase (or prepositional phrase, for frame SRL); (ii) any other category; and (iii) \nullLabel.
\end{compactitem}

In Figure~\ref{fig:macedonia}, for the span ``\textit{encouraging them}'', the constituent identity scaffold label is 1, the non-terminal label is \texttt{S$\mid$VP}, the non-terminal and parent label is \texttt{S$\mid$VP+par=PP}, and the common non-terminals label is set to \texttt{OTHER}.

\section{Semantic Role Labeling}
\label{sec:srl}

We contribute a new SRL model which contributes a strong baseline for experiments with syntactic scaffolds. The performance of this baseline itself is competitive with state-of-the-art methods \sect{sec:experiments}.

\paragraph{FrameNet.} 
In the FrameNet lexicon~\cite{Baker:98}, a \textit{frame} represents a type of event, situation, or relationship, and is associated with a set of semantic roles, called \textit{frame elements}.
A frame can be evoked by a word or phrase in a sentence, called a \textit{target}.
Each frame element of an evoked frame can then be realized in the sentence as a sentential span, called an \textit{argument} (or it can be  unrealized).
Arguments for a given frame do not overlap.

\paragraph{PropBank.}
PropBank similarly disambiguates predicates and identifies argument spans.  Targets are disambiguated to lexically specific \emph{senses} rather than shared frames, and a set of generic roles is used for all targets, reducing the argument label space by a factor of 17.
Most importantly, the arguments were annotated on top of syntactic constituents, directly coupling syntax and semantics.
A detailed example for both formalisms is provided in Figure~\ref{fig:macedonia}.

Semantic structure prediction is the task of identifying targets, labeling their frames or senses, and labeling all their argument spans in a sentence. 
Here we assume gold targets and frames, and consider only the SRL task.

Formally, a single input instance for argument identification consists of:  an $n$-word sentence $\seq{x} = \tuple{x_1, x_2, \ldots, x_n}$, a single target span $t = \tuple{t_{\text{start}}, t_{\text{end}}}$, and its evoked frame, or sense, $f$.
The argument labeling task is to produce a \term{segmentation} of the sentence: $\seq{s}=\tuple{s_1, s_2, \ldots, s_m}$ for each input $\seq{x}$.
A \term{segment} $s = \tuple{i, j, y_{i:j}}$ corresponds to a labeled span of the sentence, where the label $y_{i:j} \in \rolesForF \cup \{ \nullLabel \}$ is either a role that the span fills, or \nullLabel\ if the span does not fill any role.
In the case of PropBank, $\rolesForF$ consists of all possible roles.
The segmentation is constrained so that argument spans cover the sentence and do not overlap ($i_{k+1} = 1 + j_k$ for $s_k$; $i_1 = 1$; $j_m = n$).
Segments of length $1$ such that $i = j$ are allowed.
A separate segmentation is predicted for each target annotation in a sentence.

\subsection{Semi-Markov CRF}
\label{sec:segrnn}

In order to model the non-overlapping arguments of a given target,
we use a semi-Markov conditional random field (\term{semi-CRF}; \citealp{Sarawagi:04}).
Semi-CRFs define a conditional distribution over labeled segmentations of an input sequence, and are globally normalized.
A single target's arguments can be neatly encoded as a labeled segmentation by giving the spans in between arguments a reserved \nullLabel\ label.
Semi-Markov models are more powerful than BIO tagging schemes, which have been used successfully for PropBank SRL~\interalia{Collobert:11,Zhou:15}, because 
the semi-Markov assumption allows scoring variable-length segments, rather than fixed-length label $n$-grams as under an $(n-1)$-order Markov assumption.  
Computing the marginal likelihood with a semi-CRF can be done using dynamic programming in $O(n^2)$ time~\sect{sec:objective}.
By filtering out segments longer than $D$ tokens, this is  reduced to $O(nD)$.

Given an input $\seq{x}$, a semi-CRF  defines a conditional distribution $p(\seq{s} \mid \seq{x})$.
Every segment $s = \tuple{i,j,y_{i:j}}$ is given a real-valued score, $\psi(\tuple{i, j, y_{i:j}=r}, \seq{x}_{i:j}) = \seq{w}_{r} \cdot \spemb_{i:j}$,
where $\spemb_{i:j}$ is an embedding of the span~\sect{sec:srl-input} and $\seq{w}_{r}$ is a parameter vector corresponding to its label.
The score of the entire segmentation $\seq{s}$ is the sum of the scores of its segments: $\Psi(\seq{x}, \seq{s}) = \textstyle\sum_{k=1}^{m} \psi(s_k, \seq{x}_{i_k:j_k}).$
These scores are exponentiated and normalized to define the probability distribution.
The sum-product variant of the semi-Markov dynamic programming algorithm is used to calculate the normalization term (required during learning).  
At test time, the max-product variant returns the most probable segmentation, $\hat{\seq{s}} = \argmax{}_{\seq{s}} \Psi(\seq{s}, \seq{x})$.

The parameters of the semi-CRF are learned  to maximize a criterion related to the conditional log-likelihood of the gold-standard segments in the training corpus \sect{sec:objective}.
The learner evaluates and adjusts segment scores $\psi(s_k, \seq{x})$ for every span in the sentence, which in turn involves learning embedded representations for all spans~\sect{sec:srl-input}.

\subsection{Softmax-Margin Objective}
\label{sec:objective}

Typically CRF and semi-CRF models are trained to maximize a conditional log-likelihood objective.
In early experiments, we found that incorporating a structured cost was beneficial;
we do so by using a softmax-margin training objective \cite{Gimpel:10}, a ``cost-aware'' variant of log-likelihood:
\begin{align}
\label{eqn:softmax-margin-segrnn}
\mathcal{L}_1 =& - \displaystyle\sum_{(\seq{x}, \seq{s}^\ast) \in \mathcal{D}_1} \log \frac{\exp \Psi(\seq{s}^\ast,\seq{x})}{Z(\seq{x}, \seq{s}^\ast)}, \\
Z(\seq{x}, \seq{s}^\ast) =& \displaystyle\sum_{\seq{s}} \exp{ \{ \Psi(\seq{s}, \seq{x})  + \text{cost}(\seq{s},\seq{s}^\ast) \}}.
\end{align}
We design the cost function so that it factors by predicted span, in the same way $\Psi$ does:
\begin{align}
\text{cost}(\seq{s},\seq{s}^\ast) =&
  \displaystyle\sum_{s \in \seq{s}}{ \text{cost}(s, \seq{s}^\ast) } =
  \displaystyle\sum_{s \in \seq{s}}{ \ind{s \not\in \seq{s}^\ast} }.
\end{align}
The softmax-margin criterion, like log-likelihood, is globally normalized over all of the exponentially many possible labeled segmentations.
The following zeroth-order semi-Markov dynamic program \cite{Sarawagi:04} efficiently computes the new partition function:
\begin{align}
\alpha_j =&
\displaystyle\sum_{
	\substack{ s = \tuple{i, j, y_{i:j}} \\ j - i \leqslant D }} { \alpha_{i-1} \exp \{ \Psi(s, \seq{x}) + \text{cost}(s, \seq{s}^\ast) \}},
\end{align}
where $Z = \alpha_n$, under the base case $\alpha_0 = 1$.  

The prediction under the model can be calculated using a similar dynamic program with the following recurrence where $\gamma_0=1$:
\begin{align}
\gamma_j =
\displaystyle\max_{\substack{ s = \tuple{i, j, y_{i:j}} \\ j - i \leqslant D }} { \gamma_{i-1} \exp \Psi(s, \seq{x})}.
\end{align}
Our model formulation enforces that arguments do not overlap.
We do not enforce any other SRL constraints, such as non-repetition of core frame elements~\cite{Das:12}.

\subsection{Input Span Representation}
\label{sec:srl-input}
This section describes the neural architecture used to obtain the span embedding, $\spemb_{i:j}$, corresponding to a span $\seq{x}_{i:j}$ and the target in consideration, $t = \tuple{t_{\text{start}}, t_{\text{end}}}$.
For the scaffold task, since the syntactic treebank does not contain annotations for semantic targets, we use the last verb in the sentence as a placeholder target, wherever target features are used. 
If there are no verbs, we use the first token in the sentence as a placeholder target.
The parameters used to learn $\spemb$ are shared between the tasks.

We construct an embedding for the span using 
\begin{compactitem}
	\item $\hidden_i$ and $\hidden_j$: contextualized embeddings for the words at the span boundary~\sect{sec:boundary},
	\item $\seq{u}_{i:j}$: a span summary that pools over the contents of the span~\sect{sec:summary}, and
	\item $\seq{a}_{i:j}$: and a hand-engineered feature vector for the span~\sect{sec:features}.
\end{compactitem}

This embedding is then passed to a feedforward layer to compute the span representation, $\spemb_{i:j}$.

\subsubsection{Contextualized Token Embeddings}
\label{sec:boundary}
To obtain contextualized embeddings of each token in the input sequence, we run a bidirectional LSTM~\cite{Graves:12} with $\ell$ layers over the full input sequence. 
To indicate which token is a predicate, a linearly transformed one-hot embedding $\seq{v}$ is used, following \citet{Zhou:15} and \citet{He:17}.
The input vector representing the token at position $q$ in the sentence is the concatenation of a fixed pretrained embedding $\mathbf{x}_q$ and $\seq{v}_q$.
When given as input to the bidirectional LSTM, this yields a hidden state vector $\seq{h}_q$ representing the $q$th token in the context of the sentence.

\subsubsection{Span Summary}
\label{sec:summary}
Tokens within a span might convey different amounts of information necessary to label the span as a semantic argument.
Following \citet{Lee:17}, we use an attention mechanism~\cite{Bahdanau:14} to summarize each span.
Each contextualized token in the span is passed through a feed-forward network to obtain a weight, normalized to give $\sigma_k = \softmax{i \leqslant k \leqslant j} \seq{w}^{\text{head}} \cdot \hidden_k,$
where $\seq{w}^{\text{head}}$ is a learned parameter.
The weights $\sigma$ are then used to obtain a vector that summarizes the span, $\seq{u}_{i:j} = \textstyle \sum_{i \leqslant k \leqslant j; j -i < D} \sigma_k \cdot \hidden_k$.

\subsubsection{Span Features}
\label{sec:features}
We use the following three features for each span: 
\begin{compactitem}
    \item width of the span in tokens~\cite{Das:14}
    \item distance (in tokens) of the span from the target~\cite{Tackstrom:15}
    \item position of the span with respect to the target (\textit{before}, \textit{after}, \textit{overlap})~\cite{Tackstrom:15}
\end{compactitem}
Each of these features is encoded as a one-hot-embedding and then linearly transformed to yield a feature vector, $\seq{a}_{i:j}$.

\section{Coreference Resolution}
\label{sec:coref}
Coreference resolution is the task of determining clusters of mentions that refer to the same entity.
Formally, the input is a document $\seq{x} = x_1, x_2, \ldots, x_n$ consisting of $n$ words.
The goal is to predict a set of clusters $\seq{c} = \{c_1, c_2, \ldots\}$, where each cluster $c = \{s_1, s_2, \ldots\}$ is a set of spans and each span $s = \tuple{i, j}$ is a pair of indices such that $1 \leqslant i \leqslant j \leqslant n$.

As a baseline, we use the model of \citet{Lee:17}, which we describe  briefly in this section.
This model decomposes the prediction of coreference clusters into a series of span classification decisions.
Every span $s$ predicts an antecedent $w_s \in \mathcal{Y}(s) = \{\nullLabel, s_1, s_2, \ldots, s_m\}$. 
Labels $s_1$ to $s_m$ indicate a coreference link between $s$ and one of the $m$ spans that precede it, and $\nullLabel$ indicates that $s$ does not link to anything, either because it is not a mention or it is in a singleton cluster. 
The predicted clustering of the spans can be recovered by aggregating the predicted links.

Analogous to the SRL model~\sect{sec:srl}, every span $s$ is represented by an embedding $\spemb_s$, which is central to the model.
For each span $s$ and a potential antecedent $a \in \mathcal{Y}(s)$, pairwise coreference scores $\Psi(\spemb_s, \spemb_a, \phi(s, a))$ are computed via  feedforward networks with the span embeddings as input. $\phi(s, a)$ are pairwise discrete features encoding the distance between span $s$ and span $a$ and metadata, such as the genre and speaker information. 
We refer the reader to  \citet{Lee:17} for the details of the scoring function.

The scores from $\Psi$ are normalized over the possible antecedents $\mathcal{Y}(s)$ of each span to induce a probability distribution for every span:
\begin{align}
p(w_s = a) &= 
\softmax{a \in \mathcal{Y}(s)} \Psi(\spemb_s, \spemb_a, \phi(s, a))
\end{align}

In learning, we minimize the negative log-likelihood marginalized over the possibly correct antecedents:
\begin{align}
\mathcal{L}_1 = \displaystyle -\sum_{s \in \mathcal{D}}\log\sum_{a^* \in G(s) \cap \mathcal{Y}(s)} p(w_s = a^*)
\end{align}
\noindent where $\mathcal{D}$ is the set of spans in the training dataset, and $G(s)$ indicates the gold cluster of $s$ if it belongs to one and $\{\nullLabel\}$ otherwise.

To operate under reasonable computational requirements, inference under this model requires a two-stage beam search, which reduces the number of span pairs considered.  We refer the reader to \citet{Lee:17} for details.

\paragraph{Input span representation.}
The input span embedding, $\spemb_{s}$ for coreference resolution and its syntactic scaffold follow the definition used in \S\ref{sec:srl-input}, with the key difference of using no target features.
Since there is a complete overlap of input sentences between $\mathcal{D}_{\text{sc}}$ and $\mathcal{D}_{\text{pr}}$ as the coreference annotations are also from OntoNotes~\cite{Pradhan:12}, we  reuse the $\spemb$ for the scaffold task.
Additionally, instead of the entire document, each sentence in it is independently given as input to the bidirectional LSTMs.

\section{Results}
\label{sec:experiments}

We evaluate our models on the test set of FrameNet 1.5 for frame SRL and on the test set of OntoNotes for both PropBank SRL and coreference.
For the syntactic scaffold in each case, we use syntactic annotations from OntoNotes 5.0~\cite{Weischedel:13,Pradhan:13}.\footnote{\url{http://cemantix.org/data/ontonotes.html}}
Further details on experimental settings and datasets have been elaborated in the supplemental material.

\paragraph{Frame SRL.} 
Table \ref{tab:fn_results} shows the performance of all the scaffold models on frame SRL with respect to prior work and a semi-CRF baseline~\sect{sec:segrnn} without a syntactic scaffold.
We follow the official evaluation from the SemEval shared task for frame-semantic parsing \cite{Baker:07}.

Prior work for frame SRL has relied on predicted syntactic trees, in two different ways:
by using syntax-based rules to prune out spans of text that are unlikely to contain any frame's argument; and by using syntactic features in their statistical model \citep{Das:14,Tackstrom:15,Fitzgerald:15,Kshirsagar:15}.

The best published results on FrameNet 1.5 are due to \citet{Yang:17}.
In their sequential model (\textsc{seq}), they treat argument identification as a sequence-labeling problem using a deep bidirectional LSTM with a CRF layer.
In their relational model (\textsc{Rel}), they treat the same problem as a span classification problem.
Finally, they introduce an ensemble to integerate both models, and use an integer linear program for inference satisfying SRL constraints.
Though their model does not do any syntactic pruning, it does use syntactic features for argument identification and labeling.\footnote{\citet{Yang:17} also evaluated on the full frame-semantic parsing task, which includes frame-SRL as well as identifying frames.  
Since our frame SRL performance improves over theirs, we expect that incorporation into a full system (e.g., using their frame identification module) would lead to overall benefits as well; this experiment is left to future work.}

Notably, all prior systems for frame SRL listed in Table \ref{tab:fn_results} use a pipeline of syntax and semantics.
Our semi-CRF baseline outperforms all prior work, without any syntax.
This highlights the benefits of modeling spans  and  of global normalization.

Turning to scaffolds, even the most coarse-grained constituent identity scaffold improves the performance of our syntax-agnostic baseline.
The nonterminal and nonterminal and parent scaffolds, which use more detailed syntactic representations, improve over this.
The greatest improvements come from the scaffold model predicting common nonterminal labels (\texttt{NP} and \texttt{PP}, which are the most common syntactic categories of semantic arguments, vs.~others):  
3.6\% absolute improvement in $F_1$ measure over prior work.

Contemporaneously with this work, \citet{Peng:18} proposed a system for joint frame-semantic and semantic dependency parsing.
They report results for joint frame and argument identification, and hence cannot be directly compared in Table \ref{tab:fn_results}.
We evaluated their output for argument identification only; our semi-CRF baseline model exceeds their performance by 1 $F_1$, and our common nonterminal scaffold by 3.1 $F_1$.\footnote{This result is not reported in Table~\ref{tab:fn_results} since \citet{Peng:18} used a preprocessing which renders the test set slightly larger --- the difference we report is calculated using their test set.}

\begin{table}[tbh]
	\small
	\center
	\begin{tabulary}{\columnwidth}{@{}l   rr  r@{}}
		\toprule
		
		\textbf{Model}
		& \textbf{Prec.}
		& \textbf{Rec.}
		& \textbf{$\bm{F_1}$}\\
		
		\midrule
		
		\citet{Kshirsagar:15}
		& 66.0 & 60.4 & 63.1 \\
		
		\citet{Yang:17} (\textsc{Rel})
		& 71.8 & 57.7 
		& 64.0 \\
		
		\citet{Yang:17} (\textsc{Seq})
		& 63.4 & 66.4 
		& 64.9 \\
		
		$\dagger$\citet{Yang:17} (\textsc{All})
		& 70.2 & 60.2 
		& 65.5 \\
		
		\midrule[.03em]		
		
		Semi-CRF baseline
		& 67.8 & 66.2 & 67.0 \\ 
		\midrule
		
		~~~~~~+ constituent identity
		& 68.1 & 67.4 & 67.7 \\ 
		~~~~~~+ nonterminal and parent
		& 68.8 & 68.2 & 68.5 \\ 
		~~~~~~+ nonterminal
		& 69.4 & 68.0 & 68.7 \\ 
		~~~~~~+ common nonterminals
		& 69.2 & 69.0 & \bf{69.1} \\ 
		\bottomrule
	\end{tabulary}
	\caption{Frame SRL results on the test set of FrameNet 1.5., using gold frames. Ensembles are denoted by $\dagger$.}
	\label{tab:fn_results}
\end{table}

\begin{table}[tbh]
	\small
	\center
	\begin{tabulary}{\columnwidth}{@{}l   rr  r@{}}
		\toprule
		
		\textbf{Model}
		& \textbf{Prec.}
		& \textbf{Rec.}
		& \textbf{$\bm{F_1}$}\\
		
		\midrule
		\citet{Zhou:15}
		& -     & -    & 81.3 \\
		\citet{He:17}
		&  81.7 & 81.6 & 81.7 \\
	
        \citet{He:18} 
		&  83.9 & 73.7 & 82.1 \\
		
		\citet{Tan:18}
		&  81.9 & 83.6 & 82.7 \\

		\midrule[.03em]		
		
		Semi-CRF baseline
		&  84.8 &  81.2 &  83.0 \\ 
		\midrule

		~~~~~~+ common nonterminals
		& 85.1 & 82.6 & \bf 83.8 \\
		
		
		\bottomrule
	\end{tabulary}
	\caption{PropBank sSRL results, using gold predicates, on CoNLL 2012 test. For fair comparison, we show only non-ensembled models.}
	\label{tab:onto_results}
\end{table}

\begin{table*}[tbh]
\small
\setlength{\tabcolsep}{0.2em}
\def\arraystretch{0.95}
\centering
\begin{tabulary}{\linewidth}{l c*{14}{c}}
\toprule
\textbf{Model} & \multicolumn{4}{c}{\textbf{MUC}} &  \multicolumn{4}{c}{\textbf{$\text{B}^3$}} & \multicolumn{4}{c}{\textbf{$\text{CEAF}_{\phi_4}$}} & \multicolumn{1}{c}{\textbf{Avg. $\bm{F_1}$}}\\ 
 & Prec.  & Rec.  & $F_1$ & \; & Prec. & Rec. & $F_1$ & \;  & Prec.  & Rec.  & $F_1$ & \; &   \\
\midrule
\newcite{Wiseman:16}    & 77.5 & 69.8 & 73.4 && 66.8 & 57.0 & 61.5 && 62.1 & 53.9 & 57.7 && 64.2\\
\newcite{clark:2016a}     & 79.9 & 69.3 & 74.2 && 71.0 & 56.5 & 63.0 && 63.8 & 54.3 & 58.7 && 65.3\\ 
\newcite{Clark:16}        & 79.2 & 70.4 & 74.6 && 69.9 & 58.0 & 63.4 && 63.5 & 55.5 & 59.2 && 65.7\\ 
\cmidrule{1-14} 
\newcite{Lee:17}          & 78.4 & 73.4 & 75.8 && 68.6 & 61.8 & 65.0 && 62.7 & 59.0 & 60.8 && 67.2\\
\midrule [.03em]
~~~~~~+ common nonterminals  & 78.4 & 74.3 & 76.3 && 68.7 & 62.9 & 65.7 && 62.9 & 60.2 & 61.5 && \bf 67.8 \\
\bottomrule
\end{tabulary}

\caption{Coreference resolution results on the test set on the English CoNLL-2012 shared task. 
The average $F_1$ of MUC, $\text{B}^3$, and $\text{CEAF}_{\phi_4}$ is the main evaluation metric. 
For fair comparison, we show only non-ensembled models.
}
\label{tab:test_results}
\end{table*}

\paragraph{PropBank SRL.}
We use the OntoNotes data from the CoNLL shared task in 2012~\cite{Pradhan:13} for Propbank SRL.
Table \ref{tab:onto_results} reports results using gold predicates.

Recent competitive systems for PropBank SRL follow the approach of \citet{Zhou:15}, employing deep architectures, and forgoing the use of any syntax.
\citet{He:17} improve on those results, and in analysis experiments, show that constraints derived using syntax may further improve performance.
\citet{Tan:18} employ a similar approach but use feed-forward networks with self-attention.
\citet{He:18} use a span-based classification to jointly identify and label argument spans.

Our syntax-agnostic semi-CRF baseline model improves on prior work (excluding \texttt{ELMo}), showing again the value of global normalization in semantic structure prediction.
We obtain further improvement of 0.8 absolute $F_1$ with the best syntactic scaffold from the frame SRL task.
This indicates that a syntactic inductive bias is beneficial even when using sophisticated neural architectures.

\citet{He:18} also provide a setup where initialization was done with deep contextualized embeddings, \texttt{ELMo}~\cite{Peters:18}, resulting in 85.5 $F_1$ on the OntoNotes test set.
The improvements from \texttt{ELMo} are methodologically orthogonal to syntactic scaffolds. 

Since the datasets for learning PropBank semantics and syntactic scaffolds \textit{completely} overlap, the performance improvement cannot be attributed to a larger training corpus (or, by extension, a larger vocabulary), though that might be a factor for frame SRL.

A syntactic scaffold can match the performance of a pipeline containing carefully extracted syntactic features for semantic prediction~\cite{Swayamdipta:17}.
This, along with other recent approaches~\cite{He:17,He:18b} show that syntax remains useful, even with strong neural models for SRL.

\paragraph{Coreference.} 
We report the results on four standard scores from the  CoNLL evaluation: MUC, B$^3$ and CEAF$_{\phi_{4}}$, and their average $F_1$ in Table~\ref{tab:test_results}. 
Prior competitive coreference resolution systems \cite{Wiseman:16,clark:2016a,Clark:16} all incorporate synctactic information in a pipeline, using features and rules for mention proposals from predicted syntax. 

Our baseline is the model from ~\citet{Lee:17}, described in \S\ref{sec:coref}.
Similar to the baseline model for frame SRL, and in contrast with prior work, this model does not use any syntax.

We experiment with the best syntactic scaffold from the frame SRL task.
We used \texttt{NP}, \texttt{OTHER}, and \nullLabel as the labels for the common nonterminals scaffold here, since coreferring mentions are rarely prepositional phrases.
The syntactic scaffold outperforms the baseline by 0.6 absolute $F_1$.
Contemporaneously, \citet{Lee:18} proposed a model which takes in account higher order inference and more aggressive pruning, as well as initialization with \texttt{ELMo} embeddings, resulting in 73.0 average $F_1$.
All the above are orthogonal to our approach, and could be incorporated to yield higher gains.

 \section{Discussion}
\label{sec:analysis}

\begin{figure*}
\centering
\begin{minipage}{.45\linewidth}
    \includegraphics[width=.99\columnwidth]{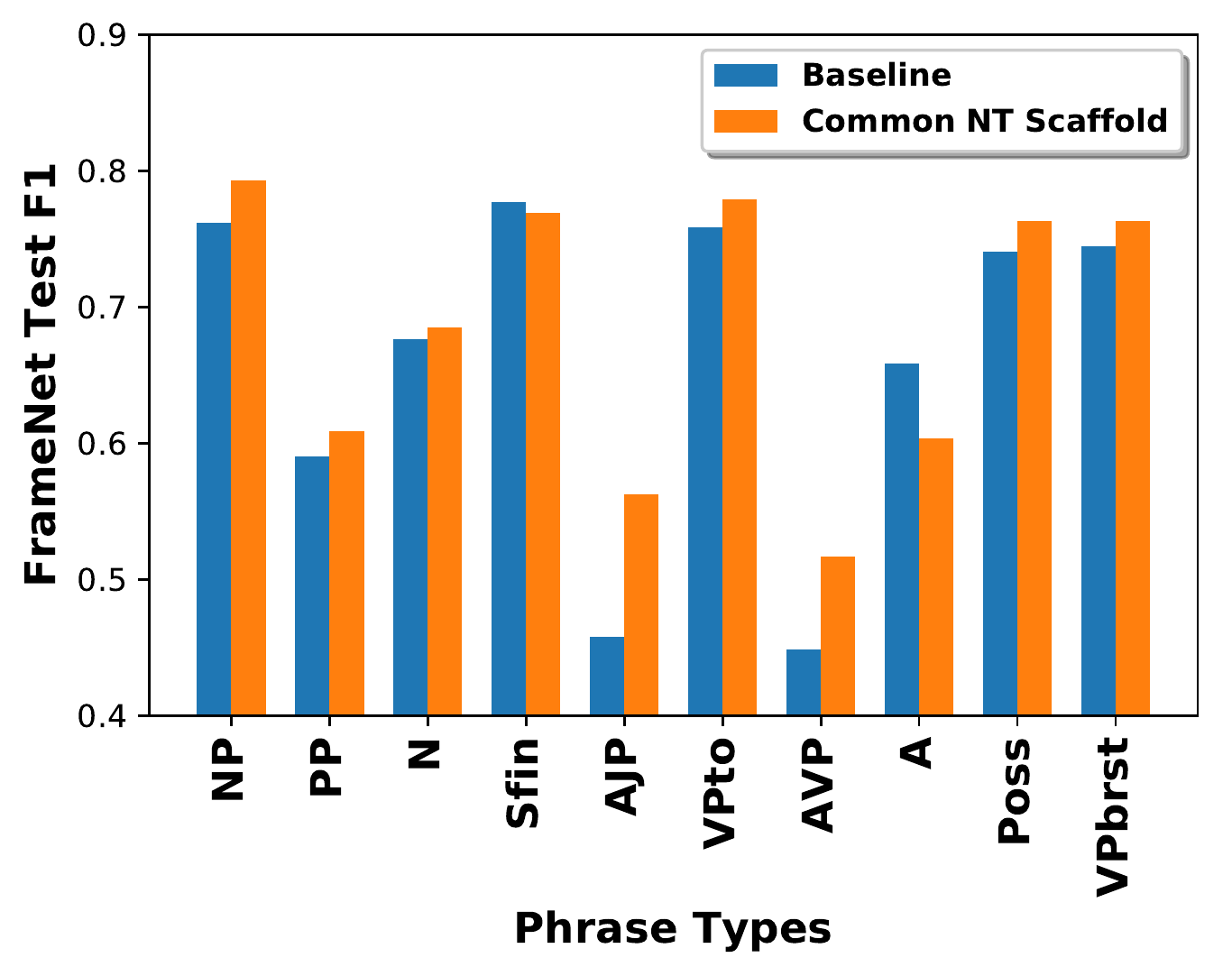}
    \caption{Performance breakdown by argument's phrase category, sorted left to right by frequency, for top ten phrase categories.}
    \label{fig:fn_phrase_types}
\end{minipage}
\hspace{.05\linewidth}
\begin{minipage}{.45\linewidth}
    \includegraphics[width=.99\columnwidth]{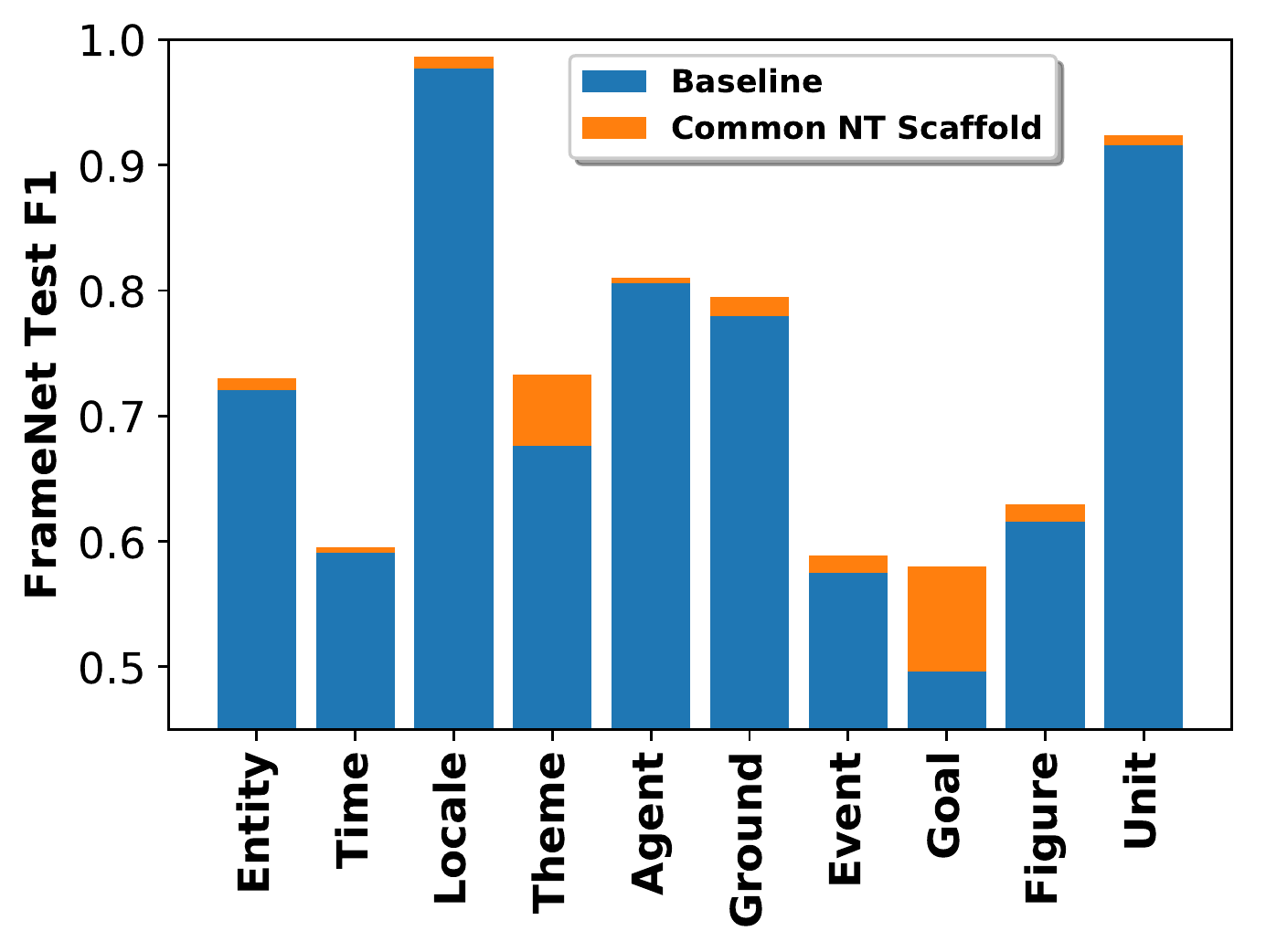}
    \caption{Performance breakdown by top ten frame element types, sorted left to right by frequency.}
    \label{fig:fn_freq_fes}
\end{minipage}
\end{figure*}

To investigate the performance of the syntactic scaffold, we focus on the frame SRL results, where we observed the greatest improvement with respect to a non-syntactic baseline.


		
		
		
		

We consider a breakdown of the performance by the syntactic phrase types of the arguments, provided in FrameNet\footnote{We used FrameNet syntactic phrase annotations for analysis only, and not in our models, since they are annotated only for the gold arguments.} in Figure~\ref{fig:fn_phrase_types}.
Not surprisingly, we observe large improvements in the common nonterminals used (\texttt{NP} and \texttt{PP}).
However, the phrase type annotations in FrameNet do not correspond exactly to the OntoNotes phrase categories.
For instance, FrameNet annotates non-maximal (\texttt{A}) and standard adjective phrases (\texttt{AJP}), while OntoNotes annotations for noun-phrases are flat, ignore the underlying adjective phrases.
This explains why the syntax-agnostic baseline is able to recover the former while the scaffold is not.

Similarly, for frequent frame elements, scaffolding improves performance across the board, as shown in Fig.~\ref{fig:fn_freq_fes}.
The largest improvements come for \texttt{Theme} and \texttt{Goal}, which are predominantly realized as noun phrases and prepositional phrases.


\section{Conclusion}
\label{sec:conclusion}

We introduced syntactic scaffolds, a multitask learning approach to incorporate syntactic bias into semantic processing tasks. 
Unlike pipelines and approaches which jointly model syntax and semantics, no explicit syntactic processing is required at runtime.
Our method improves the performance of competitive baselines for semantic role labeling on both FrameNet and PropBank, and for coreference resolution. 
While our focus was on span-based tasks, syntactic scaffolds could be applied in other settings (e.g., dependency and graph representations).
Moreover, scaffolds need not be syntactic; we can imagine, for example, \emph{semantic} scaffolds being used to improve NLP applications with limited annotated data.
It remains an open empirical question to determine the relative merits of different kinds of scaffolds and multi-task learners, and how they can be most productively combined.
Our code is publicly available at \url{https://github.com/swabhs/scaffolding}.

\section*{Acknowledgments}
We thank several members of UW-NLP, particularly Luheng He, as well as David Weiss and Emily Pitler for thoughtful discussions on prior versions of this paper.
We also thank the three anonymous reviewers for their valuable feedback.
This work was supported in part by NSF grant IIS-1562364 and by the NVIDIA Corporation through the donation of a Tesla GPU.

\bibliographystyle{acl_natbib_nourl}
\bibliography{emnlp2018}

\clearpage
\appendix

\section{Supplementary Material}
\label{sec:supplemental}
\subsection{Datasets}


We used the full-text portion of FrameNet 1.5 release\footnote{A later release, 1.7 is also available, but for ease of comparison to other published systems we report results on the earlier release.} for frame-semantic role labeling.
We use the same test set as \citet{Das:14}, and create a validation set by selecting 8 documents from the train set.
The dataset contains 3,139 train sentences with 16,621 target annotations, 387 validation sentences with 2,282 targets, and 2,420 test sentences with 4,427 targets. 
Each target from a given sentence is treated as an independent training instance.
Following \citet{Tackstrom:15}, we only use the first annotation for each target with multiple annotations.

We use the standard splits provided in OntoNotes for the CoNLL 2012 shared task.
The dataset contains 115,812 train sentences with 278,026 target annotations, 15,680 validation sentences with 38,377 targets, and 12,217 test sentences with 29,669 targets. 

We use the English coreference resolution data from the CoNLL 2012 shared task~\cite{Pradhan:12}, containing 2,802, 343 and 348 documents for train, validation, and test respectively.

\paragraph{Syntax}
OntoNotes contains 115,812 training instances for the syntactic scaffold.
There is no overlap between FrameNet and OntoNotes training data.

\subsection{Experimental Settings} 
We used GloVe embeddings~\cite{Pennington:14} for tokens in the vocabulary, with out of vocabulary words being initialized randomly.
For frame-SRL, 300 dimensional embeddings were used, and kept fixed during training.
For PropBank SRL, we used 100 dimensional embeddings which were updated during training.
A 100-dimensional embedding is learned for indicating target positions, following ~\citet{Zhou:15}.
Bidirectional LSTMs with highway connections~\cite{Srivastava:15} between 6 layers are used, each layer containing 300-dimensional hidden states.
A dropout of 0.1 is applied to the LSTMs.
The feed-forward networks are of dimension 150 and of depth 2, with rectified linear units~\cite{Nair:10}.
A dropout of 0.2 is applied to the feed-forward networks.

We limit the maximum length of spans to $D =$ 15 in FrameNet, resulting in  oracle recall of 95\% on the development set, and to 13 in Propbank, resulting in an oracle recall of 96\%.
An identical maximum span length is used for the scaffold task.

For the SRL scaffolds, we randomly sample instances from OntoNotes to match the size of the SRL data, and alternate between training an SRL batch and a scaffold batch.
In FrameNet, this amounts to downsampling OntoNotes.
For PropBank SRL, this amounts to upsampling syntactic annotations from OntoNotes, since a sentence has a single syntactic tree, but could have multiple target annotations, each of which is a training instance.

The mixing ratio, $\delta$ is set to 1.0 (tuned across \{0.1, 0.5, 1.0, 1.5\}) for frame and PropBank SRL.
We use Adam~\cite{Kingma:14} for optimization, at a learning rate of 0.001, and a minibatch of size 32.
Our dynamic program formulation for loss computation and inference under the semi-CRF is also batched.
To prevent exploding gradients, the 2-norm of the gradient is clipped to 1 before a gradient update~\cite{Graves:13}.
All models are trained for a maximum of 20 epochs, and stopped early based on dev $F_1$.

We extended the \ai ~library,\footnote{\url{http://allennlp.org/}} which is built on top of  PyTorch.\footnote{\url{http://pytorch.org/}}
Each experiment was run on a single TitanX GPU.

For the coreference model, we use the same hyperparameters and experimental settings from \citet{Lee:17}. 
The only new hyperparameter needed for scaffolding is the mixing ratio, $\delta$, which we set to 0.1 based on performance on the validation set.
\end{document}